# IMPROVING THE CHARACTER RECOGNITION EFFICIENCY OF FEED FORWARD BP NEURAL NETWORK

Amit Choudhary[1] and Rahul Rishi[2]

[1]Department of Computer Science, Maharaja Surajmal Institute, New Delhi, India
`amit.choudhary69@gmail.com`
[2]Department of Computer Science and Engineering, TITS, Bhiwani, Haryana, India
`rahulrishi@rediffmail.com`

## ABSTRACT

*This work is focused on improving the character recognition capability of feed-forward back-propagation neural network by using one, two and three hidden layers and the modified additional momentum term. 182 English letters were collected for this work and the equivalent binary matrix form of these characters was applied to the neural network as training patterns. While the network was getting trained, the connection weights were modified at each epoch of learning. For each training sample, the error surface was examined for minima by computing the gradient descent. We started the experiment by using one hidden layer and the number of hidden layers was increased up to three and it has been observed that accuracy of the network was increased with low mean square error but at the cost of training time. The recognition accuracy was improved further when modified additional momentum term was used.*

## KEYWORDS

*Character Recognition, MLP, Hidden Layers, Back-propagation, Momentum Term.*

## 1. INTRODUCTION

Off-line character recognition involves the automatic conversion of hand printed character (as an image) into letter codes which are usable within computer and text-processing applications. As compared to on-line; off-line character recognition is comparatively difficult, as different people have different handwriting styles and also the characters are extracted from documents of different intensity and background [6]. Nevertheless, limiting the range of variations in input can allow recognition process to improve.

One of the most important types of feed forward neural network is the Back Propagation Neural Network (BPNN) [12]. It is a multi-layer feed forward network using gradient-descent based delta-learning rule, commonly known as back propagation (of errors) rule. Back Propagation provides a computationally efficient method for changing the weights in a feed forward network, with differentiable activation function units, to learn a training set of input-output examples. Being a Gradient Descent Method, it minimizes the total squared error of the output computed by the net.

The network is trained by supervised learning method. The aim is to train the network to achieve a balance between the ability to respond correctly to the input characters that are used for training and the ability to provide good responses to the input that were similar. The total squared error of the output computed by network is minimized by a gradient descent method known as Back Propagation or Generalized Delta Learning Rule [1].

The experiments conducted in this paper have shown the effect on the learning and character recognition accuracy of the neural network by increasing the number of hidden layers and introducing an additional modified momentum term. Three experiments were performed. Experiment-1, Experiment-2 and Experiment-3 employed a network having one, two and three hidden layers respectively. All other experimental conditions such Learning Rate ($\eta$), Momentum Constant ($\alpha$), Activation Function, Maximum Training Epochs, Acceptable Error Level and Termination Condition were kept same for all the experiments.

The remainder of the paper is organized as follows: Section 2 briefs some related work already done by the researchers in this field. Section 3 deals with the overall system design and the various steps involved in the OCR system. Neural Network Architecture and functioning of proposed experiments are presented in section 4. Various experimental conditions for all the experiments are given in Section 5. Discussion of Results and interpretations are described in section 6. Section 7 presents the conclusion and also gives the future path for continual work in this field.

## 2. RELATED WORKS

A number of review papers on off-line handwriting recognition have been published [4, 14].In the review, Steinherz, et al. [14] commented on the importance of features extraction and selection for the recognition system to perform well. Vinciarelli [5] focused on segmentation based strategies as far as off-line handwritten word recognition is concerned. He pointed out that these strategies were suitable for small lexical only. The review presented by Koerich, et al. [4] focused on the large lexical based systems in which large number of training samples is required. In the literature, very good recognition results have been seen as far as isolated numerals or characters are concerned [7]. However, the results obtained for the segmentation and recognition of cursive handwritten words have not been satisfactory in comparison [10, 11, 13, 15].The reason for not achieving satisfactory recognition rates is the difficult nature of cursive handwriting and difficulties in the accurate segmentation and recognition of cursive and touching characters.

## 3. OCR SYSTEM DESIGN

The various steps involved in the recognition of a handwritten character are illustrated in the form of flow chart in Fig. 1.

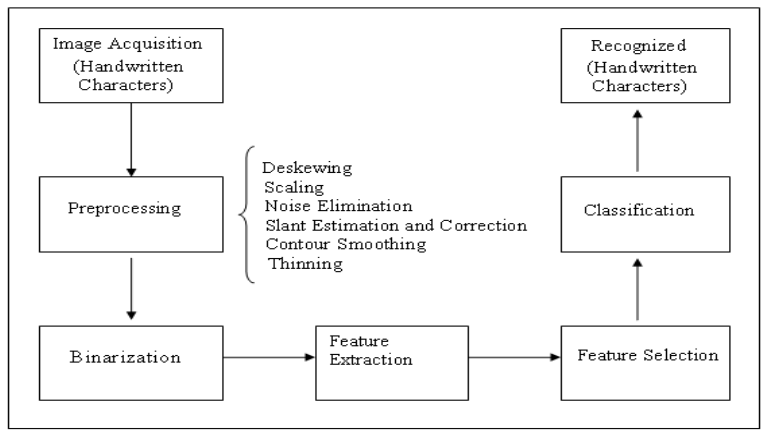

Figure 1. Typical Off-Line Character Recognition System

The steps required for typical off-line character recognition are described here in detail:-

### 3.1 Pre-processing

Pre-processing is done to remove the variability that was present in off-line handwritten characters. The pre-processing techniques that have been employed in an attempt to increase the performance of the recognition process are as follows:

*Deskewing* is used to make the base line of the handwritten word in a horizontal direction by rotating the word in a suitable direction by a suitable angle. Some examples of techniques for correcting slope are described by Brown and Ganapathy [9].

*Scaling* sometimes may be necessary to produce characters of relative size

*Noise* can be removed by comparing the character image by a threshold [6].

*Slant estimation and correction* is achieved by analysis of the slanted vertical projections at various angles [2].

*Contour Smoothing* is a technique to remove contour noise which is introduced in the form of bumps and holes due to the process of slant correction.

*Thinning* is a process in which the skeleton of the character image is used to normalize the stroke width.

### 3.2 Binarization

All hand printed characters were scanned into grayscale images. Each character image was traced vertically after converting the grayscale image into binary matrix [3, 8]. The threshold parameter along with the grayscale image was made an input to the binarization program designed in MATLAB. The output was a binary matrix which represented the image shown in Fig. 2(c).Every character was first converted into a binary matrix and then resized to 8 X 6 matrixes as shown in Fig. 2(c) and reshaped to a binary matrix of size 48 X 1 which is made as an input to the neural network for learning and testing. Binary matrix representation of character 'A' can be defined as in Fig. 2(c). The resized characters were clubbed together in a matrix of size 48 X 26 to form a sample [2]. In the sample, each column corresponds to an English alphabet which was resized into 48 X 1 input vector.

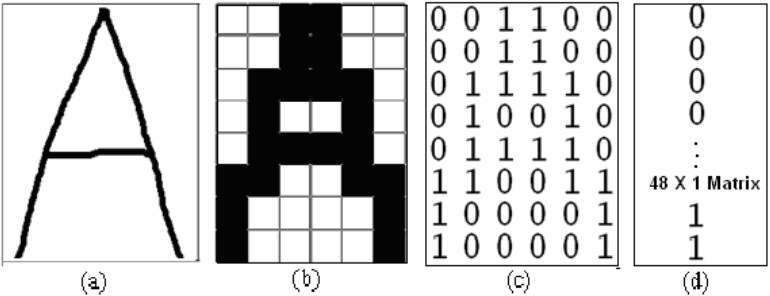

Figure 2 (a) Grayscale image of character 'A' (b) Binary representation of character 'A'; (c) Binary matrix representation and (d) Reshaped sample of character 'A'.

For sample creation, 182 characters were collected from 35 people. After pre-processing, 5 samples were considered for training such that each sample was consisting of 26 characters (A-Z) and 2 samples were considered for testing the recognition accuracy of the network.

### 3.3 Feature Extraction and Selection

The derived information can be general features, which were evaluated to ease further processing.

*3.4 Classification*

Classification is the final stage of our OCR system design. This is the stage where an automated system declares that the inputted character belongs to a particular category. The classifier here we have used is a feed forward back propagation neural network.

# 4. NEURAL NETWORK ARCHITECTURE USED IN THE RECOGNITION PROCESS

To accomplish the task of character classification and input-output mapping, the multi-layer feed forward artificial neural network was considered with nonlinear differentiable function 'tansig' in all processing units of output and hidden layers. The neurons in the input layer have linear activation function. The number of output units corresponds to the number of distinct classes in the pattern classification. A method has been developed, so that network can be trained to capture the mapping implicitly in the set of input output pattern pair collected during an experiment and simultaneously expected to modal the unknown system to function from which the predictions can be made for the new or untrained set of data [8, 12]. The possible output pattern class would be approximately an interpolated version of the output pattern class corresponding to the input learning pattern close to the given test input pattern. This method involved the back propagation learning rule based on the principle of gradient descent along the error surface in the negative direction.

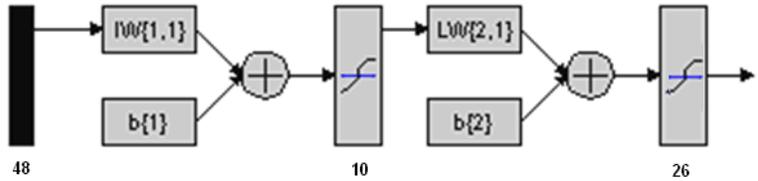

Figure 3. Feed forward neural network with one hidden layer.

The network has 48 input neurons that are equivalent to the input character's size as we have resized every character into a binary matrix of size 8 X 6. The number of neurons in the output layer was 26 because there are 26 English alphabets. The number of hidden neurons is directly proportional to the system resources. The bigger the number more the resources are required. The number of neurons in a hidden layer was kept 10 for optimal results.

The output of the network can be determined as

$$y_k = f(\sum_{i=1}^{n} Z_i W_{ik})$$

where $f$ is the output function,

$Z_i$ is the output of hidden layer and

$W_{ik}$ is the connection strength between neurons of hidden and output layer.

and , also for hidden layer's processing unit output ;

$$Z_i = f(\sum_{j=1}^{n} V_{ij} X_j)$$

where $X_j$ is the output of input layer and $V_{ij}$ is the weight between input and hidden layer.

The LMS error between the desired and actual output of the network

$$E = 0.5 \sum_k [t_k - y_k]^2$$

where $t_k$ is desired output

The error minimization can be shown as;

$$\partial E \Big/ \partial W_{ik} = [t_k - y_k] f'(y_k) z_i$$

Weights modifications on the hidden layer can be defined as;

$$\Delta V_{ij} = \partial E \Big/ \partial V_{ij} = \sum_k \partial k * (\partial y_k \Big/ \partial v_{ij})[t_k - y_k] f'(y_k) z_i$$

[Let, $\partial k = [t_k - y_k] f'(y_k)$]

and we have

$$\Delta V_{ij} = \sum_k \partial k * W_{ik} f'(z_i)$$

Thus the weight updates for output unit can be represented as;

$$W_{ik}(t+1) = W_{ik}(t) + \eta \Delta W_{ik}(t) + \alpha \Delta W_{ik}(t-1)$$

where

$W_{ik}(t)$ is the state of weight matrix at iteration t

$W_{ik}(t+1)$ is the state of weight matrix at next iteration

$W_{ik}(t-1)$ is the state of weight matrix at previous iteration.

$\Delta W_{ik}(t)$ is current change/ modification in weight matrix and

$\alpha$ is standard momentum variable to accelerate learning process. This variable depends on the learning rate of the network. As the network yields the set learning rate the momentum variable tends to accelerate the process.

The network is made to learn the behaviour with this Gradient Descent. For next trial the gradient momentum term is modified by adding one more term i.e. the second momentum term.

$$W_{ik}(t+1) = W_{ik}(t) + \eta \Delta W_{ik}(t) + \alpha \Delta W_{ik}(t-1) + \beta \Delta W_{ik}(t-2)$$

When the weight update with equation is computed sequentially, leads to the following benefits:

1. Formed as a sum of the current (descent) gradient direction and a scaled version of the previous correction.
2. Faster learning process
3. Weight modification is based on the behaviour learned from latest three iterations instead of two.

The neural network was exposed to 5 different samples as achieved in Section 3. Actual output of the network was obtained by "COMPET" function [3] and is a binary matrix of size 26×26 because each character has 26×1 output vector. First 26×1 column stores the first character's recognition output, the following column will be for next character and so on for 26 characters.

For each character the 26×1 vector will contain value '1' at only one place. For example character 'A' if correctly recognized, will result in [1, 0, 0, 0 …all …0] and character 'B' will result in [0, 1, 0, 0 … all …0]. The difference between the desired and actual output was calculated for each cycle and the weights were adjusted during back-propagation. The process continued till the network converged to the allowable or acceptable error.

## 5. LEARNING PARAMETERS

The various parameters and their respective values used in the learning process of all the three experiments with one, two and three hidden layers are shown in Table I.

Table I:  Experimental Conditions of the neural  network

| PARAMETER | VALUE | |
|---|---|---|
| **Input Layer** | | |
| No. of Input Neurons | 48 | |
| Transfer / Activation Function | Linear | |
| **Hidden Layer** | | |
| No. of  Neurons | 10 | |
| Transfer / Activation Function | TanSig | |
| **Output Layer** | | |
| No. of Output Neurons | 26 | |
| Transfer / Activation Function | TanSig | |
| Learning Rule | Momentum | |
| Learning Constant | 0.01 | |
| Acceptable Error (MSE) | 0.001 | |
| Momentum Term ($\alpha$) | 0.90 | |
| Modified Momentum Term ($\beta$) | 0.05 | |
| Maximum Epochs | 2000 | |
| Termination Conditions ($N_{HL}$) | Based on minimum Mean Square Error or maximum number of epochs allowed | |
| Initial Weights and bias term values | Randomly generated values between 0 and 1 | |
| Number of Hidden Layers ($N_{HL}$) | Experiment-1 | $N_{HL} =1$ |
| | Experiment-2 | $N_{HL} =2$ |
| | Experiment-3 | $N_{HL} =3$ |

## 6. DISCUSSION OF RESULTS AND INTERPRETATIONS

The system was simulated using a feed forward neural network system that consisted of 48 neurons in input layer, 10 neurons in hidden layer and 26 output neurons. The characters were resized into 8×6 binary matrixes and were exposed to 48 input neurons. The 26 output neurons correspond to 26 upper case letters of English alphabet. The network having one hidden layer was used for Experiment-1 and in Experiment-2 and Experiment-3; the process was repeated for the network having two and three hidden layers each having 10 hidden neurons. In structured sections, the experiments and their outcomes at each stage are described.

*Gradient Computation*

The gradient descent is the characteristic of error surface. If the surface is not smooth, the gradient calculated will be a large number and this will give a poor indication of the "true error correction path". On the other hand, if the surface is relatively smooth, the gradient value will be a smaller one. Hence the smaller gradient is always the desirable one. For each trial of learning, the computed values of gradient descent are shown in Table II.

Table II: Comparison of gradient values of the network for all the three experiments with standard and modified momentum term.

| Sample | Experiment-1($N_{HL}$=1) | | Experiment-2($N_{HL}$=2) | | Experiment-3($N_{HL}$=3) | |
|---|---|---|---|---|---|---|
| | Gradient1 (Classical Method) | Gradient2 (Modified Method) | Gradient1 (Classical Method) | Gradient2 (Modified Method) | Gradient1 (Classical Method) | Gradient2 (Modified Method) |
| Sample1 | 1981400 | 2029280 | 1419834 | 1300263 | 1217348 | 1114823 |
| Sample2 | 5792000 | 2021500 | 3714695 | 3502623 | 2984628 | 2132849 |
| Sample3 | 7018400 | 1723310 | 5834838 | 5459346 | 4629835 | 4017892 |
| Sample4 | 1173900 | 1043094 | 6157572 | 4914614 | 6276419 | 1222478 |
| Sample5 | 6226319 | 3189781 | 6317917 | 5718393 | 5186437 | 3048174 |

It has been observed in Table II that in Experiment 2 using MLP with two hidden layers, the gradient values are much smaller than in MLP with one hidden layer used in Experiment 1. It is also observed that as the number of hidden layer is further increased by one, the gradient value is found to be least as shown in Experiment 3 in Table II. It is clear from all the three experiments that the gradient values are further reduced when modified momentum term is introduced in the weight update process.

*Number of Epochs*

The number of epochs in the learning process of the network is represented in Table III.

Table III: Comparison of training epochs between the learning trails for all the three experiments

| Sample | Experiment-1($N_{HL}$=1) | | Experiment-2($N_{HL}$=2) | | Experiment-3($N_{HL}$=3) | |
|---|---|---|---|---|---|---|
| | Epoch1 (Standard Momentum Term) | Epoch2 (Modified Momentum Term) | Epoch1 (Standard Momentum Term) | Epoch2 (Modified Momentum Term) | Epoch1 (Standard Momentum Term) | Epoch2 (Modified Momentum Term) |
| Sample1 | 186 | 173 | 521 | 479 | 909 | 881 |
| Sample2 | 347 | 321 | 623 | 598 | 1104 | 1023 |
| Sample3 | 551 | 529 | 717 | 680 | 1334 | 1252 |
| Sample4 | 695 | 663 | 832 | 778 | 1391 | 1428 |
| Sample5 | 811 | 759 | 960 | 972 | 1569 | 1463 |

In Table III, Epoch1 and Epoch2 represent the number of network iterations for a particular sample when presented to the neural network having standard and modified momentum terms respectively in all the three experiments. It is clear that small number of epochs is sufficient to train a network when we use one hidden layer. As the number of hidden layers is increased, the number of epochs also increases as observed in Experiment 3 in Table III. We can say that the network converges slowly when a large number of hidden layers are used in the experiment and the network converges rapidly when the second momentum term is introduced. Although, the network with three hidden layers requires more time during learning, the gradient values are found to be quiet low as shown earlier in Table II. Hence, the error surface will be smooth and the network's probability of getting stuck in the local minima will be low.

*Error estimation*

The network performance achieved is shown in Table IV. For all the three experiments with one, two and three hidden layers, first column in each experiment represents the error present in the network trained with classical weight update method and second column represents the error present in the network trained with modified weight update method. It is evident that the error is reduced when the number of hidden layers is increased and is further reduced when the modified momentum term is used in the network during weight update mechanism. In other words, we can conclude that with the increase in the number of hidden layers, there is an increase in probability of converging the network before the number of training epochs reaches it maximum allowed count.

Table IV: Error level attained by the neural network

| Sample | Experiment-1($N_{HL}$=1) | | Experiment-2($N_{HL}$=2) | | Experiment-3($N_{HL}$=3) | |
|---|---|---|---|---|---|---|
| | Error With Classical Weight Update Method | Error With Modified Weight Update Method | Error With Classical Weight Update Method | Error With Modified Weight Update Method | Error With Classical Weight Update Method | Error With Modified Weight Update Method |
| Sample1 | 0.00006534 | 0.000059171 | 0.000023139 | 0.000023581 | 0.000012144 | 0.000005317 |
| Sample2 | 0.000568387 | 0.000492472 | 0.000374023 | 0.000329581 | 0.000024913 | 0.000005955 |
| Sample3 | 0.000831155 | 0.000618926 | 0.000550854 | 0.000517252 | 0.000049734 | 0.000008634 |
| Sample4 | 0.000912383 | 0.000637404 | 0.000834808 | 0.000751754 | 0.000058315 | 0.000031475 |
| Sample5 | 0.004875740 | 0.003585913 | 0.001218150 | 0.001397148 | 0.000094612 | 0.000182618 |

*Testing*

The character recognition accuracy with testing samples executed in all the three networks with Standard Momentum Term and Modified Momentum Term are shown in Table V. The networks were tested with two samples. These samples were new to all the three networks because they were never trained with these samples. The recognition rates for these samples are shown in Table V.

It has been observed that in Experiment 2 using MLP with two hidden layers, the recognition rates are better than MLP with one hidden layer used in Experiment 1. It is also observed that as the number of hidden layers is further increased by one, the recognition accuracy is found to be best as shown in Experiment 3 in Table V. It is clear from all the three experiments that the recognition percentage is further improved when modified momentum term is introduced in the weight update process.

Table V: Character recognition accuracy

| (Sample) Number of characters in test sample | Experiment-1($N_{HL}$=1) | | Experiment-2($N_{HL}$=2) | | Experiment-3($N_{HL}$=3) | |
|---|---|---|---|---|---|---|
| | Standard Momentum Term | Modified Momentum Term | Standard Momentum Term | Modified Momentum Term | Standard Momentum Term | Modified Momentum Term |
| | Correctly Recognized (%age) | Correctly Recognized (%age) | Correctly Recognized (%age) | Correctly Recognized (%age) | Correctly Recognized (%age) | Correctly Recognized (%age) |
| 26 | 17 (65.38%) | 22 (84.61%) | 20 (76.92%) | 22 (84.61%) | 23 (88.46%) | 24 (92.30%) |
| 26 | 20 (76.92%) | 23 (88.46%) | 21 (80.76%) | 23 (88.46%) | 22 (84.61%) | 24 (92.30%) |

When the networks using standard momentum term and having one, two and three hidden layers are being trained with Sample 1, the profiles of MSE plot for the training epochs are drawn in Fig. 4, Fig. 5 and Fig. 6 respectively.

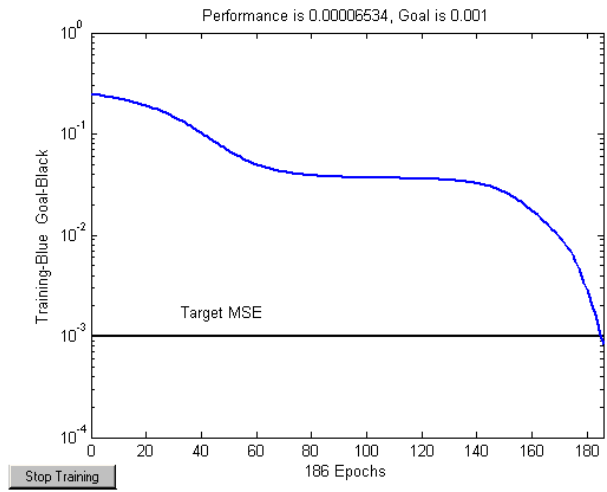

Figure 4. MSE plot for the network with one hidden layer and standard momentum term

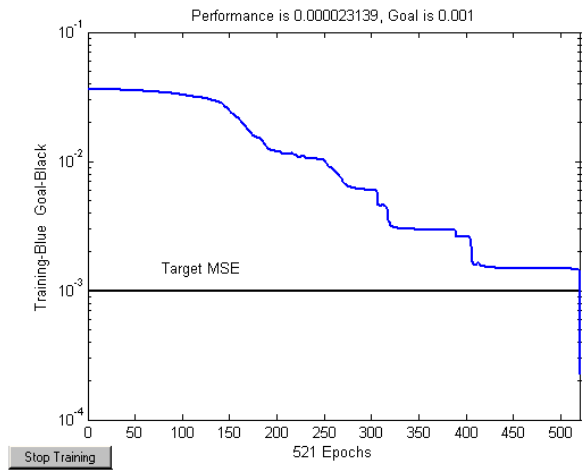

Figure 5. MSE plot for the network with two hidden layers and standard momentum term

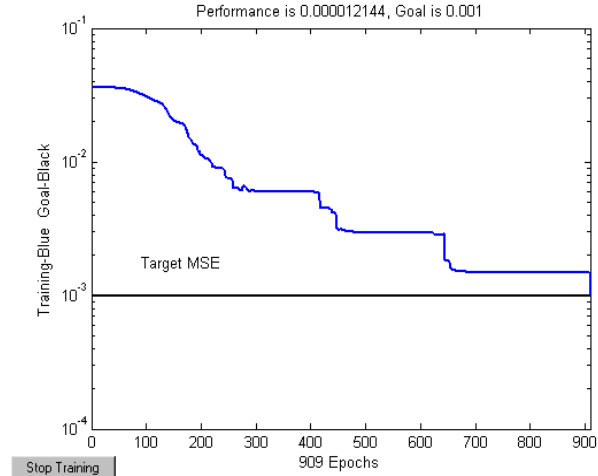

Figure 6. MSE plot for the network with three hidden layers and standard momentum term

As the number of hidden layers is increased, the network will converge slowly. After strong analysis, three relationships between the number of hidden layers, number of epochs and MSE are established.

$$N_{HL} \; \alpha \; N_E \tag{1}$$

where $N_{HL}$ is the number of hidden layers and $N_E$ is the number of epochs.

The number of training epochs is inversely proportional to the minimum MSE.

$$N_E \quad \alpha \quad \frac{1}{MSE} \tag{2}$$

The number of hidden layers is inversely proportional to the minimum MSE.

$$N_{HL} \; \alpha \quad \frac{1}{MSE}$$

$$\tag{3}$$

where MSE is the mean square error.

## 7. CONCLUSION AND FUTURE SCOPE

The proposed method for the handwritten character recognition using the descent gradient approach and modified momentum term yielded the remarkable enhancement in the performance. As shown in Table-II, the results of all the three experiments for the different sample of characters represent that the smaller gradient values are achieved in case of modified momentum weight update mechanism and gradient values are found to be least when three hidden layers were used in the network. Smaller the gradient values, smoother will be the error surface and the probability that the neural network will get stuck in the local minima will be the least. Smaller gradient values indicated that the error correction was downy and accurate.

This paper has introduced an additional momentum term in the weight modification process. This additional momentum term accelerates the process of convergence and the network shows better performance. It is clear from Table-V that the recognition accuracy is best in Experiment-3 where MLP with three hidden layers uses modified momentum term for updating its weights during error back propagation. Eq.1 implies that the number of hidden layers is proportional to the number of epochs. This means that as the number of hidden layers is increased, the training process of the network slows down and is indicated by the increase in the number of training epochs. The number of epochs required to train a network are reduced when the second momentum term is introduced during weight update mechanism as shown in Table-III. However, Eq.3 implies that the training of the network is more accurate if more hidden layers are used and the accuracy is further improved when the modified momentum term is used during weight update mechanism as shown in Table-V. This accuracy is achieved at the cost of network training time as indicated by Eq.2.

The training process of the network is improved when a second momentum term is introduced. So if the accuracy of the results is a critical factor for an application, then the network with modified momentum term and having many hidden layers should be used but if time is a critical factor then the network with modified momentum term and having single hidden layer (with sufficient number of hidden neurons) should be used.

Due to the back-propagation of error element in Multilayer Perceptron (MLP), it frequently suffers from the problem of Local-Minima; hence the samples may not converge. The network may get trapped in local minima even though there is a much deeper minimum nearby. Nevertheless, more work needs to be done especially on the test for more complex handwritten characters. The proposed work can be carried out to recognize English words of different character lengths after proper segmentation of the words into isolated character images

Thus it can be concluded that more complex neural networks with modified momentum term and having many hidden layers can be used for soft real time systems where performance is more critical.

**Authors:**

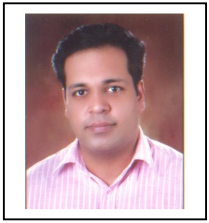

Amit Choudhary is currently working as an Assistant Professor in the Department of Computer Science at Maharaja Surajmal Institute, New Delhi, India for the last 9 years. He has done MCA, M.Tech and M.Phil in Computer science and is pursing his doctoral degree in Computer Science and Engineering from M. D. University, Rohtak, India. His research interest is focused on Machine Learning, Pattern Recognition and Artificial Intelligence. He has many international publications to his credit.

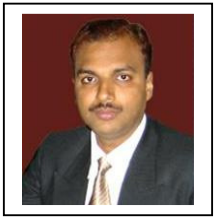

Dr. Rahul Rishi is currently working as an Associate Professor and Head of Computer Science & Engineering Department, TITS, Bhiwani, Haryana, India for the last 14 years. He has done B.Tech, M.Tech and Ph.D. in Computer Science & Engineering. He has published more than thirty five research papers in International Conferences and International Journals of repute. His current research activities pertain to Fuzzy Relational Databases, Soft Computing, Artificial Intelligence and Data Mining.